\documentclass{article}

     \PassOptionsToPackage{numbers, compress}{natbib}

     \usepackage[preprint]{neurips_2020}

\usepackage[utf8]{inputenc} 
\usepackage[T1]{fontenc}    
\usepackage{hyperref}       
\usepackage{url}            
\usepackage{booktabs}       
\usepackage{amsfonts}       
\usepackage{nicefrac}       
\usepackage{microtype}      
\usepackage[pdftex]{graphicx}
\usepackage{multirow}
\usepackage{wrapfig}
\usepackage{subfigure}
\usepackage{amsmath}
\usepackage{caption}
\title{Spatial Transformer Point Convolution}

\author{%
  Yuan Fang$^1$, Chunyan Xu$^1$, Zhen Cui$^1$, Yuan Zong$^2$, and Jian Yang$^1$\\
  $^1$Nanjing University of Science and Technology, Jiangsu, China\\
  $^2$Southeast University, Jiangsu, China\\
  \texttt{\{fangyuan, cyx, zhen.cui, csjyang\}}@njust.edu.cn \\
  \texttt{xhzongyuan@seu.edu.cn}\\
}

\usepackage{amsmath,amssymb} 
\usepackage{color}
\usepackage{graphicx}
\usepackage{amsthm}
\usepackage{bm}
\usepackage{algorithm}
\usepackage{algorithmic}
\usepackage{cases}
\usepackage{subfigure}
\usepackage{multirow}
\usepackage{enumerate}
\usepackage{etoolbox}
\usepackage{amsfonts} 
\usepackage{amssymb} 
\usepackage{boxedminipage} 
\usepackage{tabularx}
\usepackage{url} 
\usepackage{hyperref}

\newcommand{\MyMapTemplatePrefix}[4]{\expandafter#1\csname#3#4\endcsname{#2{#4}}}
\newcommand{\MyMapTemplatePrefixNew}[5]{\expandafter#1\csname#4#5\endcsname{#2{#3{#5}}}}
\forcsvlist{\MyMapTemplatePrefix {\def} {\mathbf} {}} {A,B,C,D,E,F,G,H,I,J,K,L,M,N,O,P,Q,R,S,T,U,V,W,X,Y,Z}
\forcsvlist{\MyMapTemplatePrefix {\def} {\mathbf} {}} {a,b,c,d,e,f,g,h,l,m,n,o,p,q,r,s,t,u,v,w,x,y,z,1,0}
\forcsvlist{\MyMapTemplatePrefix {\def} {\widetilde} {wt}} {A,B,C,D,E,F,G,H,I,J,K,L,M,N,O,P,Q,R,S,T,U,V,W,X,Y,Z}
\forcsvlist{\MyMapTemplatePrefix {\def} {\widetilde} {wt}} {a,b,c,d,e,f,g,h,i,j,k,l,m,n,o,p,q,r,s,t,u,v,w,x,y,z} 
\forcsvlist{\MyMapTemplatePrefixNew {\def} {\widetilde}{\mathbf} {tb}} {A,B,C,D,E,F,G,H,I,J,K,L,M,N,O,P,Q,R,S,T,U,V,W,X,Y,Z}
\forcsvlist{\MyMapTemplatePrefixNew {\def} {\widetilde}{\mathbf} {tb}} {a,b,c,d,e,f,g,h,i,j,k,l,m,n,o,p,q,r,s,t,u,v,w,x,y,z}
\forcsvlist{\MyMapTemplatePrefix {\def} {\widehat} {wh}} {A,B,C,D,E,F,G,H,I,J,K,L,M,N,O,P,Q,R,S,T,U,V,W,X,Y,Z}
\forcsvlist{\MyMapTemplatePrefix {\def} {\widehat} {wh}} {a,b,c,d,e,f,g,h,i,j,k,l,m,n,o,p,q,r,s,t,u,v,w,x,y,z}
\forcsvlist{\MyMapTemplatePrefixNew {\def} {\widehat}{\mathbf} {hb}} {A,B,C,D,E,F,G,H,I,J,K,L,M,N,O,P,Q,R,S,T,U,V,W,X,Y,Z}
\forcsvlist{\MyMapTemplatePrefixNew {\def} {\widehat}{\mathbf} {hb}} {a,b,c,d,e,f,g,h,i,j,k,l,m,n,o,p,q,r,s,t,u,v,w,x,y,z}
\forcsvlist{\MyMapTemplatePrefix {\def} {\mathcal}{mc}} {A,B,C,D,E,F,G,H,I,J,L,M,N,O,P,Q,R,S,T,U,V,W,X,Y,Z} 
\forcsvlist{\MyMapTemplatePrefix {\def} {\mathbb} {mb}} {A,B,C,D,E,F,G,H,I,J,K,L,M,N,O,P,Q,R,S,T,U,V,W,X,Y,Z}
\forcsvlist{\MyMapTemplatePrefix {\DeclareMathOperator} {} {} } {tr,diag,sgn}
\def\tp{^\intercal}

\begin{document}

\maketitle

\begin{abstract}
	Point clouds are unstructured and unordered in the embedded 3D space. In order to produce consistent responses under different permutation layouts, most existing methods aggregate local spatial points through maximum or summation operation. But such an aggregation essentially belongs to the isotropic filtering on all operated points therein, which tends to lose the information of geometric structures. In this paper, we propose a spatial transformer point convolution (STPC) method to achieve anisotropic convolution filtering on point clouds. To capture and represent implicit geometric structures, we specifically introduce spatial direction dictionary to learn those latent geometric components. To better encode unordered neighbor points, we design sparse deformer to transform them into the canonical ordered dictionary space by using direction dictionary learning. In the transformed space, the standard image-like convolution can be leveraged to generate anisotropic filtering, which is more robust to express those finer variances of local regions. Dictionary learning and encoding processes are encapsulated into a network module and jointly learnt in an end-to-end manner. Extensive experiments on several public datasets (including S3DIS, Semantic3D, SemanticKITTI) demonstrate the effectiveness of our proposed method in point clouds semantic segmentation task.	
	
\end{abstract}

\vspace{-0.2cm}
\section{Introduction} \vspace{-0.2cm}

%
%
%

Recent advances in 3D hardware sensors (e.g., 3D scanners, LiDARs and RGB-D cameras) have improved the accessibility of point clouds, which can intuitively reflect the spatial geometric information of objects in 3D space. 
Recently, 3D semantic segmentation of point clouds, which aims to label each point with a corresponding category information, has drawn increasing attention due to its broad applicability to the fields of general environment perception. 
It enables various higher-level potential applications, including autonomous driving, human-computer interaction, virtual reality and robotics. 

With the significant progress of deep learning on grid-shaped images/videos~\cite{NIPS2012_imagenet,resnet}, recent studies began to explore how to apply convolutional neural networks (CNN) on irregular-structured point clouds~\cite{qi2017pointnet,qi2017pointnet++}. 
For example, some previous methods converted point clouds data to regular data representation by employing multi-view images~\cite{su2015multi,lawin2017deep} or voxels~\cite{maturana2015voxnet,wu20153d,qi2016volumetric}, and then CNN-like operations (e.g., 2D CNN and 3D CNN) can be performed on unstructured point clouds. 
Although these methods have made some progress, the projection and voxelization steps inevitably lead to information loss of original 3D data. 
Furthermore, as a pioneering work, PointNet~\cite{qi2017pointnet} was proposed to directly process point clouds  without intermediate data conversion. It employed the shared multi-layer perceptron (MLP) to learn features of each point and then a symmetric function is used to aggregate global features. This network can achieve permutation invariance for unordered 3D point clouds, but cannot well consider the local geometric information between points (including direction, shape and topological structure).
Subsequently, several dedicated neural modules have been proposed to aggregate local geometric information of point clouds for further boosting the semantic segmentation performance, such as Pointnet++~\cite{qi2017pointnet++}, RS-CNN~\cite{liu2019relation}, 
Geo-CNN~\cite{yang2019modeling}, PointWeb~\cite{zhao2019pointweb} and ShellNet~\cite{zhang2019shellnet}. 
%
For example, PointNet++ adopted a hierarchical neural network to process a set of points sampled in a metric space~\cite{qi2017pointnet++};  RS-CNN defined the low-level relation as a compact feature vector with 10 channels, then contextual shape-aware representation was learned for all points~\cite{liu2019relation}.

However, most of these approaches always aggregate/mix these geometric information together to explore the relationship between a point and its neighbor points. Such aggregated methods essentially belong to the isotropic filtering on all operated points therein, which tends to lose the information of 3D geometric structures.

In this paper, we propose a novel spatial transformer point convolution (STPC) method to achieve anisotropic convolution filtering on point clouds, and finally boost the performance of 3D semantic segmentation task.
Given a 3D point clouds sample, our proposed STPC framework can predict the point-wise categorization in an end-to-end fashion.
Inspired by the classic bag-of-words model~\cite{bagoftheword1,bagoftheword}, we specifically build a spatial direction dictionary to represent these latent spatial directions/geometric components of local point clouds data.
By virtue of the dictionary, these unordered and unstructured neighbor points can be coordinated into the canonical dictionary space by using sparse direction encoding.
Subsequently, in the transformed dictionary space, a typical image-like convolution operation can be applied to perform the anisotropic filtering process for robustly capturing these finer geometric structure information of point clouds.
Both dictionary learning and encoding processes are encapsuled into a network module, and the entire spatial transformer point convolution is also jointed to optimize in an end-to-end neural network.
The proposed STPC framework can better capture these subtle geometric structure information, especially for those finer variances of local regions in 3D point clouds.

We summarize the main contributions as three folds:
\begin{itemize}
	\vspace{-0.15cm}\item We propose a novel spatial transformer point convolution (STPC) framework to deal with semantic segmentation on point cloud data, which successfully performs anisotropic convolution filtering on unstructured data like the standard convolution on images.
	\vspace{-0.15cm}\item We propose a direction dictionary induced spatial direction encoding method, which transforms unordered neighbor points into a latent atom-coordinated system and further well encode those subtle structure variances of local neighbor points.
	\vspace{-0.15cm}\item Comprehensive evaluations on three point cloud datasets (including S3DIS, Semantic3D, SemanticKITTI)  demonstrate the superiority of our proposed STPC when compared with other state-of-the-art methods in the point cloud semantic segmentation problem.
\end{itemize}\vspace{-0.3cm}
\section{Related Work}\vspace{-0.2cm}
In this section, we focus on point clouds data analysis and briefly review three types of existing deep learning methods according to their underlying technologies.

\textbf{Multiview-based convolution methods:} Some existing methods \cite{su2015multi,lawin2017deep,feng2018gvcnn} project point clouds into a set of renderings from different viewpoints and then the standard convolutional neural network can be applied. MVCNN\cite{su2015multi} projected a 3D object into multiple views and extracted the corresponding view-wise features, then simply max-pools multi-view features into a global descriptor for accurate object recognition. Felix et al.\cite{lawin2017deep} projected a 3D point clouds onto 2D planes from multiple virtual camera views. Then, a multi-stream FCN was used to predict pixel-wise scores on synthetic images. The final semantic label of each point was obtained by fusing the reprojected scores over different views. However, these view-based convolution methods lost spatial information during projection, and can not fully exploit the underlying geometric and structural information. Also, the performance of multi-view semantic segmentation methods is sensitive to viewpoint selection and occlusions.
%

\textbf{Voxelization-based convolution methods:} Volumetric representation naturally preserves the neighborhood structure of 3D point clouds. Its regular data format allows direct application of standard 3D convolutions~\cite{maturana2015voxnet,wu20153d,tchapmi2017segcloud,huang2016point,le2018pointgrid}. VoxNet~\cite{maturana2015voxnet} introduced a volumetric occupancy network to achieve robust 3D object recognition. 3D ShapeNet~\cite{wu20153d} was proposed to represent 3D shape by a probability distribution of binary variables on voxel grids. To alleviate the problem of computation and memory footprint growing cubically with the resolution, Kd-trees~\cite{klokov2017escape} and octrees~\cite{riegler2017octnet,wang2017o-cnn} were introduced to efficiently model point clouds. However, the voxelization step inherently causes discretization artifacts and information loss. It is non-trivial to select an appropriate grid resolution in practice.

\textbf{Point-based methods.} The point-based method directly takes raw point clouds as input without converting them to other formats. As a pioneering work, PointNet~\cite{qi2017pointnet} used shared multi-layer perceptron to learn features of each point and then a symmetric function was used to aggregate global features. This network can achieve permutation invariance for unordered 3D point clouds but the local geometric information between points can not be captured. Subsequently,
neighbouring feature pooling methods~\cite{qi2017pointnet++,zhao2019pointweb,zhang2019shellnet}, graph message passing methods~\cite{wang2018local,wang2019graph,wang2019dynamic}, kernel-based convolution methods~\cite{atzmon2018point,groh2018flex-convolution,hua2018pointwise,thomas2019kpconv}, and attention-based aggregation methods~\cite{xie2018attentional,yang2019modeling} were proposed to learn per-point local features. Although these methods have shown promising results, they did not implement the anisotropic filtering on point clouds.
To generate anisotropic filtering on point clouds, PointCNN~\cite{li2018pointcnn} learnt a $\mathcal{X}$-transformation from the input points to weight the input features associated with the points and permute the points into a latent and potential canonical order. TangentConv~\cite{tatarchenko2018tangent} was proposed by projecting local neighbor points to local tangent planes and processing them with 2D convolutions. A-CNN~\cite{komarichev2019cnn} can order neighboring in a
clockwise/counterclockwise manner on a tangent plane and then apply annular convolutions. FPConv~\cite{lin2020fpconv} performed a local flattening by learning a weight map to softly project surrounding points onto a 2D grid. 
Our STPC is different from the above anisotropic methods in three folds: i) introduces a flexible spatial direction dictionary as the latent coordinate system (each atom may be viewed as a coordinate direction), ii) uses the inspirit of sparse coding and transformer to sparsely encode neighbor points into the dictionary system, and iii) filter kernels are defined on latent coordinate directions for anisotropic convolution.

\vspace{-0.2cm}
\section{Our Approach}\vspace{-0.2cm}

\subsection{Overview}

The standard convolution on images/videos is conditioned to grid shape region, and performs anisotropic filtering in order to encode subtle texture appearances. The point clouds data are often embedded in an unstructured and non-regular 3D space, and thus the standard convolution kernels are difficult to be defined on point clouds data due to its irregular property. 
Most existing convolution methods on point clouds ~\cite{qi2017pointnet,qi2017pointnet++} take the sum-/max-aggregation on local neighbor points without considering the local space topology of points therein. 
As a contrast, we attempt to construct an \textit{anisotropic filtering} on 3D point clouds to better encode these finer characteristics of points in each local 3D spatial region. To this end, we introduce spatial direction dictionaries to coordinate these nearest points into latent normal spaces, where each atom of the dictionary implicitly defines a spatial direction. 
The previous filter operated on all nearest points can be untied to adapt to different atoms, which naturally results to anisotropic convolution filtering analogous to the standard convolution. This anisotropic filtering can be named as \textit{spatial transformer point convolution (STPC)}. 

Given one reference point $p_i$, we can sample the $K$ nearest neighbor points as a set $\mcN(p_i)=\{p_k|k=1,\cdots, K\}$. Obviously, each point has 3D coordinate value, denoted as $\p_k=(x_k,y_k,z_k)\tp$\footnote{Here the lowercase $p$ denotes one general point, while the bold-type $\p$ denotes its spatial x-y-z coordinates. }. Point clouds often carry some additional attributes, e.g., RGB color or intermediate learnt features. We denote the attributes of point $p_{i}$ as a vector $\x_i$. Next, we perform an anisotropic
filtering on the point set $\mcN(p_i)$ to produce a robust response of the reference point $p_i$. Below we define one-time spatial transformer point convolution process, which can be encapsulated into an STPC module as one layer. Formally,
\begin{align}
\x_i^{l+1}& \longleftarrow \psi_\text{conv}([\tbx_1^l,\tbx_2^l,\cdots,\tbx_M^l]),\label{eqn:x_ilp1}\\
\tbx_m^l &\longleftarrow \mcT_{\text{f}} (\{(\alpha_{km}^l,\mcF_\text{f}^l(\p_k,\x_k^l))| p_k\in\mcN(p_i)\}),\quad m=1,\cdots M,\label{eqn:tbx_ml}\\
\bm{\alpha}_k^l:&[\alpha^l_{k1},\alpha^l_{k2},\cdots, \alpha^l_{kM}]\longleftarrow\mcT_{\text{s}}(\A^l,\mcF_\text{s}(\p_k)),\quad p_k\in\mcN(p_i), \label{eqn:alpha_jl}\\
\A^l&:[\a_1^{l},\a_2^{l},\cdots,\a_M^{l}]\longleftarrow \mcF_\text{a}(\a_1^{l-1},\a_2^{l-1},\cdots,\a_M^{l-1}),\label{eqn:A_l}
\end{align}
where $l$ is the layer number of neural network, $M$ is the number of atoms $\{\a_1,\a_2,\cdots,\a_M\}$ in the spatial direction dictionary $\A$, $\bm{\alpha}_k$ is the encoding coefficient w.r.t the point $p_k$, $\tbx_m$ is the accumulated component at the $m$-th atom for all neighbor points of $p_i$, $\x_i^{l+1}$ is the anisotropic convolution response of point $p_i$. In the above formulas,
\begin{itemize}
	\item Direction dictionary leaning $\mcF_\text{a}$. Due to spatial encoding, this dictionary should emphasize spatial information (source from points positions), rather than attribute information. Meantime, the dictionary should be dynamic and depended on prior states by accompanying with stacked modules.
	\item Spatial direction encoding $\mcT_\text{s}$. The point $p_k$ is represented with the atoms of the direction dictionary by using sparse coding or nearest neighbor reconstruction. The representation coefficients describe the latent coordinate information of this point in the dictionary space. The spatial transformation $\mcF_\text{s}$ is used to extract spatial features from 3D coordinates.
	\item Spatial transformer $\mcT_\text{f}$. The neighbor points set $\mcN(p_i)$ is projected onto atom points, and forms a regular coordinate system defined by $M$ atoms. The feature transformation $\mcF_\text{f}$ is used to learn point features/attributes.  
	\item Anisotropic convolution (i.e., STPC) $\psi_\text{conv}$. As the dictionary implicitly defines a regular coordinate system, the spatial transformed features $\{\tbx_m|m=1,\cdots, M\}$ just corresponds to the atoms therein. Hence, we could perform different filtering for each of them, which really works like the standard convolution on different relative spatial positions.
\end{itemize}

\begin{figure}[!t]
	\centering	
	\includegraphics[width=1.0\linewidth]{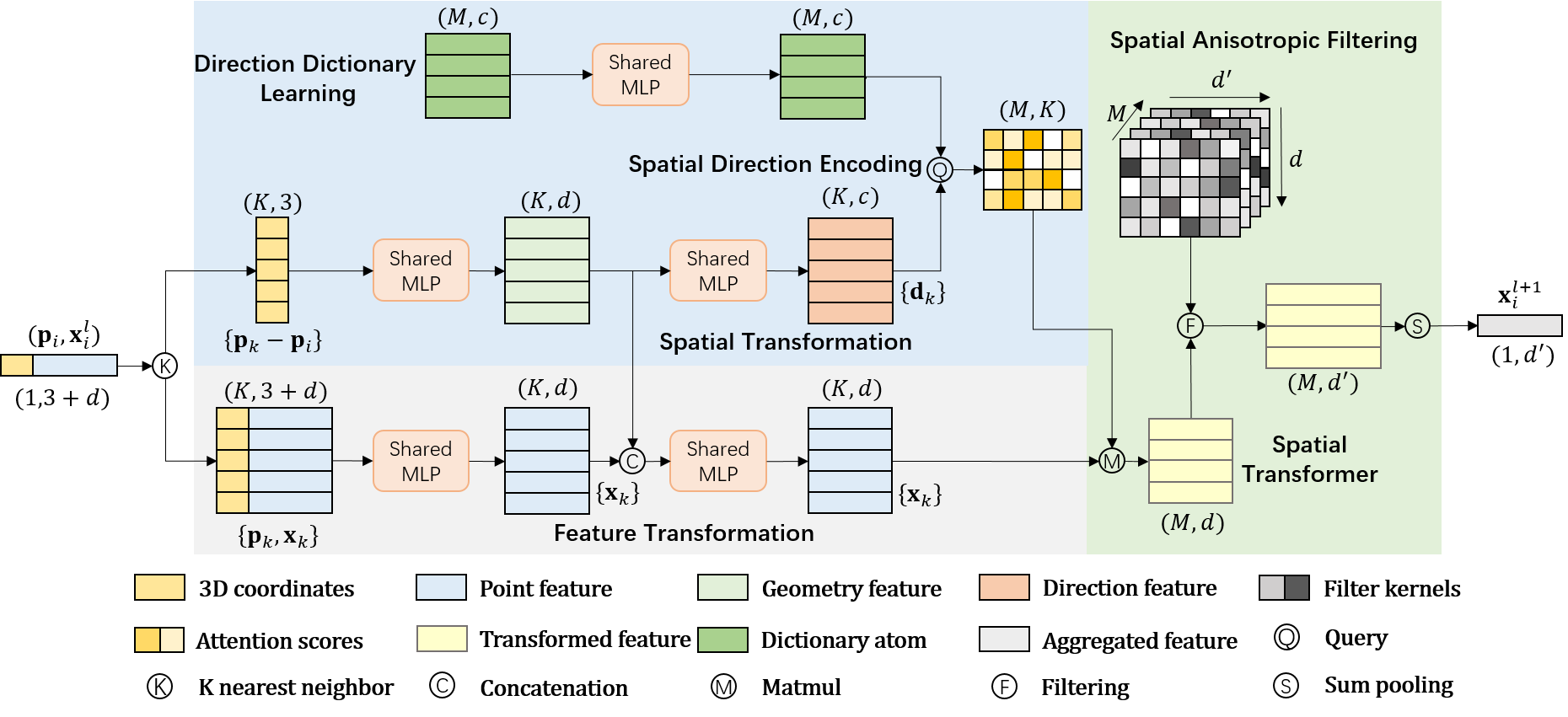}
	\caption{The overview of spatial transformer point convolution (STPC) module. Here the direction dictionary learning, spatial direction encoding, spatial anisotropic filtering can be encapsulated into the STPC module, and jointly learnt in an end-to-end fashion. The STPC module can finally achieve an anisotropic convolution filtering on point clouds for capturing those finer variances of local region. 	
	} \vspace{-0.6cm}
	\label{conv}
\end{figure}
The above convolution process can be framed into a STPC network module as shown in Fig.~\ref{conv}, whose implementation details are introduced in section~\ref{aaa}, ~\ref{bbb} and ~\ref{ccc}. 

\vspace{-0.2cm}
\subsection{Direction Dictionary Learning}\label{aaa} \vspace{-0.2cm}
In order to produce consistent responses under different permutation layouts, most existing methods mix these geometric information together to explore the relationship between a point and its neighbor points. 
Intuitively, these geometric/direction information of point clouds lie in a dense space with several latent components/directions. 
Inspired by the idea of bag-of-words, here we learn a spatial direction dictionary in the $c$-dimensional space, which can be then used to sparsely quantize local neighborhood structure  of point clouds. 
Specifically, we build a spatial direction dictionary $\A=[\a_{1}, \a_{2}, \dots, \a_{M}]$ with $M$ atoms through random initialization, where each atom $\a_{m} \in \mathbb{R}^c $ refer to a spatial direction information. 
Here the initialized dictionary can be noted as $\A^0=[\a_{1}^0, \a_{2}^0, \dots, \a_{M}^0]$. 
%
Due to the sparsity of point clouds, a certain number of atoms are sufficient to encode each neighborhood region, and the redundancy and expensive computing costs can be also avoided in practice. 
With the prior states of direction dictionary in previous layer, we can dynamically optimize/update it while satisfying the constrain of atom decorrelation. 
\begin{equation}
\begin{aligned}
&\min \sum_{p=1}^{M}\sum_{q\neq{p}}^{M}\mathcal{S}_{(\a_p^l,\a_q^l)}, \qquad \text{s.t.} \ \ \A^l=\mcF_\text{a}( \A^{l-1}), 
\label{eq:eps4} 
\end{aligned}
\end{equation}
where $\A^{l}: [\a_1^{l},\a_2^{l},\cdots,\a_M^{l}] $ and $\A^{l-1} : [\a_1^{l-1},\a_2^{l-1},\cdots,\a_M^{l-1}]$ denote the states of direction dictionary in the $l$-th and $l-1$-th layers, the dictionary learning function $\mcF_\text{a}$ is implemented by a shared multi-layer perceptron. 
In the direction dictionary learning process, we use the cosine similarity to measure the similarity between each of them.
\begin{equation}\label{eq:eps5} 
\mathcal{S}_{(\a_p^l,\a_q^l)}=\frac{\a_p^l\cdot\a_q^l}{\|\a_p^l\|\|\a_q^l\|}, \qquad p,q=1,2,\dots,M,
\end{equation}
%
where the greater the cosine similarity $\mathcal{S}_{(\a_p^l,\a_q^l)}$ is,  the more similar the spatial directions between $\a_p^l$ and $\a_q^l$ are. The constraint is that the atoms of the spatial direction dictionary are discrete and not similar between them. 
We can iteratively learn the spatial direction dictionary by minimizing the above constraint, and then the atoms can discretely represent different spatial  directions in the high-dimensional dictionary space.

\vspace{-0.2cm}
\subsection{Spatial Direction Encoding}\label{bbb}\vspace{-0.2cm}
For a single point, its x-y-z coordinates can not fully explore the local geometric information with its neighbor points, which is important for the semantic segmentation task. Several recent works\cite{yang2019modeling,liu2019relation,hu2019randla} have explicitly embedded the low-level elements of neighbor points to encode local features. 
These above methods focus on expressing local structure relationships by employing multiple low-level elements, but the distinguishable spatial direction information of each point would be drowned out. 
%
Therefore, we introduce a spatial direction encoding method to capture and represent implicit geometric structures and distinguishable direction information. 
For the $i$-th point, its spatial direction information can be represented as $[\d_1, \d_2,\dots, \d_K]\in\mathbb{R}^{K\times c}$, and each of them can be encoded as: 
\begin{equation}\label{eq:eps3} 
\begin{aligned}
\d_{k} = \mcF_\text{s}(\p_i-\p_{k}), \ \ \ \  p_k \in \mcN(p_{i}),
\end{aligned}
\end{equation}
where the point set $\{p_{k}|k=1, 2, \dots, K\}$ refers to the $K$ nearest neighbor points of the point $p_{i}$,
$c$ is the feature dimension of $\d_{k}$,
and the spatial transformation $\mcF_\text{s}$ is performed by the shared multi-layer perceptron for extracting the spatial direction information.

In the $l$-th convolution layer, we can encode the local spatial information of point $p_i$ by employing the direction dictionary $\A^{l}$ and spatial direction information $[\d_k | k=1, 2, \dots, K]$. 
For one nearest point $p_k$ of the point $p_i$, we can encode it with the atoms of the  direction dictionary by using sparse coding. 
Analogous to the self-attention mechanism~\cite{vaswani2017attention} in the NLP field, $\d_{k}$ and $\a_m^{l}$ refer to a query and one dictionary atom, respectively.  
For describing the latent coordinate information of the point $p_k$, we can get the representation coefficient by computing the correlation between spatial direction information $\d_{k}$ and the atom $\a_m^{l}$. 
Specifically, we use the cosine similarity as Eqn. (\ref{eq:eps5}) to measure the correlation between them and an exponential function is used to make the representation coefficients more sparse. 
The  spatial encoding process $\mathcal{T}_{\text{s}}$ can be expressed as computing the direction-dependent relationship between $p_k$ and each atom $\a_m^{l} \in \A^l$:
\begin{equation}\label{eq:eps8} 
\alpha_{km}^l=\frac{\exp\left(\mathcal{S}_{(\d_{k},\a_m^l)}\right)}{\sum_{m=1}^{M}\exp\left(\mathcal{S}_{(\d_{k},\a_m^l)}\right)}, 
\end{equation}
where $\alpha_{km}^l$ may be viewed as the attention score/coefficient of the point $p_k$ w.r.t the atom $\a_m$. Thus the spatial information of the point $p_k$ in the dictionary space is encoded as $\bm{\alpha}_k^l=[\alpha^l_{k1},\alpha^l_{k2},\cdots, \alpha^l_{kM}]$. 
Besides, the study on sparse coding~\cite{yang2009linear} demonstrates that the sparsity of representation coefficients is important to overcomplete dictionaries. Hence, we discard those trivial coefficients in $\bm{\alpha}$ by setting a small threshold value (i.e., $\tau$ = 0.01), which results into sparse representation in practice.
\vspace{-0.2cm}
\subsection{Spatial Anisotropic Filtering}\label{ccc}\vspace{-0.2cm}

To encode the unordered local information of the point $p_i$, we introduce a spatial transformer $\mathcal{T}_{\text{f}}$ for projecting the neighbor points set $\mathcal{N}(p_i)$ into a canonical order dictionary space. After obtaining the attention score, the feature of all neighbor points $p_k\in\mcN(p_i)$ can be transformed into the canonical ordered dictionary space.
\begin{equation}\label{eq:eps9} 
\begin{aligned}
\tbx_m^l &= \mcT_{\text{f}} (\{(\alpha_{jm}^l,\mcF_\text{f}^l(\p_k,\x_k^l))| p_k\in\mcN(p_i)\}) \\
 &= \sum_{k=1}^{K} \alpha_{km}^l 
\mcF_\text{f}^l(\p_k,\x_k^l), \quad m=1,\cdots M,
\end{aligned}
\end{equation}
where the feature transformation $\mcF_\text{f}^l(\p_k,\x_k^l)$ refers to the feature extracting process by employing the shared multi-layer perceptron, the x-y-z coordinates $\p_k$ and the geometry-aware features $\x_k^l$ of the point $p_{k}$ are used as inputs. 
Thus, any disordered neighborhood can be transformed into the canonical ordered dictionary space, and then we can perform an anisotropic convolution filtering $\psi_\text{conv}$ to produce a robust response of the point $p_i$. 
\begin{equation}
\begin{aligned}
\x_i^{l+1} &= \psi_\text{conv}([\tbx_1^l,\tbx_2^l,\cdots,\tbx_M^l]) \\
           & =\w_1*\tbx_1^l + \w_2*\tbx_2^l + \dots + \w_M*\tbx_M^l, \\
\end{aligned}
\end{equation}
where $[\tbx_1^l,\tbx_2^l,\cdots,\tbx_M^l]$ denotes the spatial transformed features with the ordered structure in the dictionary space, $[\w_1, \w_2, \dots, \w_M]$ refers to the learnt weights for performing anisotropic convolution. 

\vspace{-0.2cm}
\section{Experiments}\vspace{-0.2cm}

\subsection{Experiment Setting}

To demonstrate the efficiency of our proposed STPC method, we conduct semantic segmentation experiments on three large-scale point clouds datasets, including S3DIS~\cite{armeni2017joint}, Semantic3D~\cite{hackel2017semantic3d} and SemanticKITTI~\cite{behley2019semantickitti}. 
Given an point clouds as input, the widely-used encoder-decoder architecture is used to predict the point-wise label in an end-to-end way. 
In the encoding stage, we stack five STPC modules to extract per-point features, each of which is followed by a random sampling operation~\cite{hu2019randla} to reduce the size of the point clouds. 
In the decoding stage, five upsampling layers are stacked to propagate features by employing the nearest point interpolation technique. 
Meanwhile, skip connection is applied to concatenate the upsampled features with the intermediate features produced by encoding layers. 
To predict the categories of input point clouds, we stack several fully-connected layers on this responses, and use the cross-entropy loss during training.
Follow the same protocols as in~\cite{hu2019randla}, we split piont cloud data into training set and testing set, and adopt the same original features as inputs (3D coordinates and color information for S3DIS and Semantic3D, and only 3D coordinates for SemanticKITTI). We adopt the mean IoU (mIoU), mean class Accuracy (mAcc) and Overall Accuracy (OA) over the total classes as the standard metrics.
The learning rate is initialized at 0.01 and decreases by 5\% after each epoch. The number of nearest points K is set as 16. 
As in RandLA-Net~\cite{hu2019randla}, we use the Adam optimizer with default parameters and the same data processing methods. Our learnt spatial direction dictionary contains 25 atoms (i.e., $M=25$), the dimension of each atom is 16 (i.e., $c=16$), each of which represents a different spatial direction. 
For smooth updates, the learning rate of direction dictionary learning is set to 0.01 times the global network. 
All models in our experiment are trained and tested based on an NVIDIA RTX2080Ti GPU.

\begin{table}[!t]
	\caption{Comparison of semantic segmentation performance on the S3DIS dataset (Area 5)}
	\label{s3dis_Area_5}
	\centering 
	\resizebox{\textwidth}{!}{
		\begin{tabular}{rcccccccccccccccc}
			\toprule
			Method&OA(\%)&mAcc(\%)&mIoU(\%)&ceil.&floor&wall&beam&col.&wind.&door&table&chair&sofa&book.&board&clut. \\
			\midrule
			PointNet~\cite{qi2017pointnet}&-&49.0&41.1&88.8&97.3&69.8&0.1&3.9&46.3&10.8&58.9&52.6&5.9&40.3&26.4&33.2\\
			SegCloud~\cite{tchapmi2017segcloud}&-&48.9&57.4&90.1&96.1&69.9&0.0&18.4&38.4&23.1&70.4&75.9&40.9&58.4&13.0&41.6\\
			TangentConv~\cite{tatarchenko2018tangent}&-&62.2&52.6&90.5&97.7&74.0&0.0&20.7&39.0&31.3&77.5&69.4&57.3&38.5&48.8&39.8\\
			SPGraph~\cite{landrieu2018large}&86.4&66.5&58.0&89.4&96.9&78.1&0.0&\textbf{42.8}&48.9&61.6&84.7&75.4&69.8&52.6&2.1&52.2\\
			PointCNN~\cite{li2018pointcnn}&85.9&63.9&57.2&92.3&98.2&79.4&0.0&17.6&22.8&62.1&74.4&80.6&31.7&66.7&62.1&56.8\\
			PointConv~\cite{wu2019pointconv}&85.4&64.7&58.3&92.8&96.3&77.0&0.0&18.2&47.7&54.3&\textbf{87.9}&72.8&61.6&65.9&33.9&49.3\\
			PAT~\cite{yang2019modeling}&-&70.8&60.1&\textbf{93.0}&\textbf{98.5}&72.3&\textbf{1.0}&41.5&\textbf{85.1}&38.2&57.7&83.6&48.1&67.0&61.3&33.6\\
			PointWeb~\cite{zhao2019pointweb}&87.0&66.7&60.3&92.0&\textbf{98.5}&79.4&0.0&21.1&59.7&34.8&76.3&88.3&46.9&69.3&64.9&52.5\\
			KPConv rigid~\cite{thomas2019kpconv}&-&70.9&65.4&92.6&97.3&81.4&0.0&16.5&54.5&\textbf{69.5}&80.2&90.1&66.4&74.6&63.7&58.1\\
			KPConv deform~\cite{thomas2019kpconv}&-&72.8&\textbf{67.1}&92.8&97.3&\textbf{82.4}&0.0&23.9&58.0&69.0&81.5&\textbf{91.0}&75.4&\textbf{75.3}&\textbf{66.7}&\textbf{58.9}\\
			Our STPC&\textbf{88.5}&\textbf{75.3}&66.6 &91.7 &96.7 &82.1 &0.0 &37.3 &64.7 &52.6 &79.7 &89.2 &\textbf{76.3} &72.5 &66.5 &56.1 \\
			\bottomrule  
	\end{tabular}} 
\end{table}
\begin{table}[!t] \vspace{-0.4cm}
	\caption{Comparison of semantic segmentation performance on the Semantic3D dataset (reduced-8)}
	\label{Semantic3d}
	\centering
	\resizebox{\textwidth}{!}{
		\begin{tabular}{rcccccccccc}
			\toprule
			Method&mIoU(\%)&OA(\%)&man-made&natural&high veg.&low veg.&buildings&hard scape&scanning art.&cars\\
			\midrule
			SegCloud~\cite{tchapmi2017segcloud} &61.3 &88.1 &83.9 &66.0 &86.0 &40.5 &91.1 &30.9 &27.5 &64.3\\
			ShellNet~\cite{zhang2019shellnet} &69.3 &93.2 &96.3 &90.4 &83.9 &41.0 &94.2 &34.7 &43.9 &70.2\\
			GACNet~\cite{wang2019graph} &70.8 &91.9 &86.4 &77.7 &\textbf{88.5} &\textbf{60.6} &94.2 &37.3 &43.5 &77.8\\
			SPGraph~\cite{landrieu2018large} &73.2 &94.0 &97.4 &92.6 &87.9 &44.0 &83.2 &31.0 &63.5 &76.2\\
			KPConv rigid~\cite{thomas2019kpconv} &74.6 &92.9 &90.9 &82.2 &84.2 &47.9 &94.9 &40.0 &\textbf{77.3} &\textbf{79.7}\\
			RGNet~\cite{truong2019fast} &74.7 &94.5 &97.5 &\textbf{93.0} &88.1	&48.1 &94.6	&36.2 &72.0 &68.0\\
			RandLA-Net~\cite{hu2019randla} &\textbf{77.4} &\textbf{94.8} &95.6 &91.4 &86.6 &51.5 &\textbf{95.7} &\textbf{51.5} &69.8 &76.8\\
			Our STPC &76.2 &\textbf{94.8} &\textbf{97.9} &\textbf{93.0} &88.1 &49.3 &95.4 &47.3 &59.0 &79.6\\
			\bottomrule
	\end{tabular}} \vspace{-0.2cm}
\end{table}
\vspace{-0.2cm}
\subsection{Results and Comparisons}\vspace{-0.2cm}

\textbf{S3DIS dataset}\cite{armeni2017joint}: Table~\ref{s3dis_Area_5} shows the performance of our STPC method and comparisons with several state-of-the-arts on OA, mAcc and mIoU metrics. our STPC method can significantly outperform four baselines: 2.1\% over SPGraph~\cite{landrieu2018large}, 2.6\% over PointCNN~\cite{li2018pointcnn}, 3.1\% over PointConv~\cite{wu2019pointconv} and 1.5\% over PointWeb~\cite{zhao2019pointweb} in term of OA score. 
When compared with these existing point clouds semantic segmentation approaches, our method in term of mAcc score also achieves the best result, representing improvement of 26.3\%, 13.1\%, 11.4\%, 4.5\%, and 2.5\% over PointNet~\cite{qi2017pointnet},  TangentConv~\cite{tatarchenko2018tangent},
 PointCNN~\cite{li2018pointcnn}, PAT~\cite{yang2019modeling}, KPConv deform~\cite{thomas2019kpconv}. 
The proposed STPC method can also obtain a better performance than these exiting methods on mIoU score, except for KPConv deform method~\cite{thomas2019kpconv}. 
Furthermore, we also report the mIoU scores for each class in Table~\ref{s3dis_Area_5}. 
Generally, when compared with these baselines, our STPC method shows the comparable performance, e.g, 76.3\% \textit{vs} 75.4\%~\cite{thomas2019kpconv} for sofa, 89.2\% \textit{vs} 91.0~\cite{thomas2019kpconv} for chair, 66.5\% \textit{vs} 64.9~\cite{zhao2019pointweb} for board, and 82.1\% \textit{vs} 79.4\%~\cite{li2018pointcnn} for wall. 
It demonstrates that our STPC method performs very well on predicting the point-wise labels by considering the anisotropic convolution filtering technology on 3D point clouds data. 


\textbf{Semantic3D dataset}~\cite{hackel2017semantic3d}: 
We report the results of comparisons between the baselines and our STPC method 
on the Semantic3D dataset (reduced-8). As shown in Table~\ref{Semantic3d}, comparisons with previous
methods~\cite{tchapmi2017segcloud,zhang2019shellnet,thomas2019kpconv,truong2019fast} demonstrate that our proposed
STPC obtains the best performance, achieving improvements
of 0.3\% over RGNet~\cite{truong2019fast}, 1.9\% over  KPConv rigid~\cite{thomas2019kpconv}, 1.6\% over ShellNet~\cite{zhang2019shellnet} and 6.7\% over SegCloud~\cite{tchapmi2017segcloud} in term of OA score. 
The STPC achieves 76.2\% in term of OA score, which can outperform most existing methods~\cite{tchapmi2017segcloud,zhang2019shellnet,thomas2019kpconv,truong2019fast} , except the RandLA-Net method~\cite{hu2019randla}.  
For example, when compared with the classic SPGraph~\cite{landrieu2018large}, we can boost the semantic segmentation performance by 0.8 and 3.0\% in terms of OA and mIoU scores, respectively. 
Furthermore, when predicting the labels for each category, the mIoU scores for man-made and natural can achieve the best performance with 97.9\% and 93.0\% segmentation results. 
This indicates that our SPTC can better capture these finer local features of point clouds and then improve the discriminative capability of semantic segmentation network. 

\begin{figure*}
	\begin{minipage}[b]{0.6\linewidth}
		\centering
		\includegraphics[width=1\linewidth]{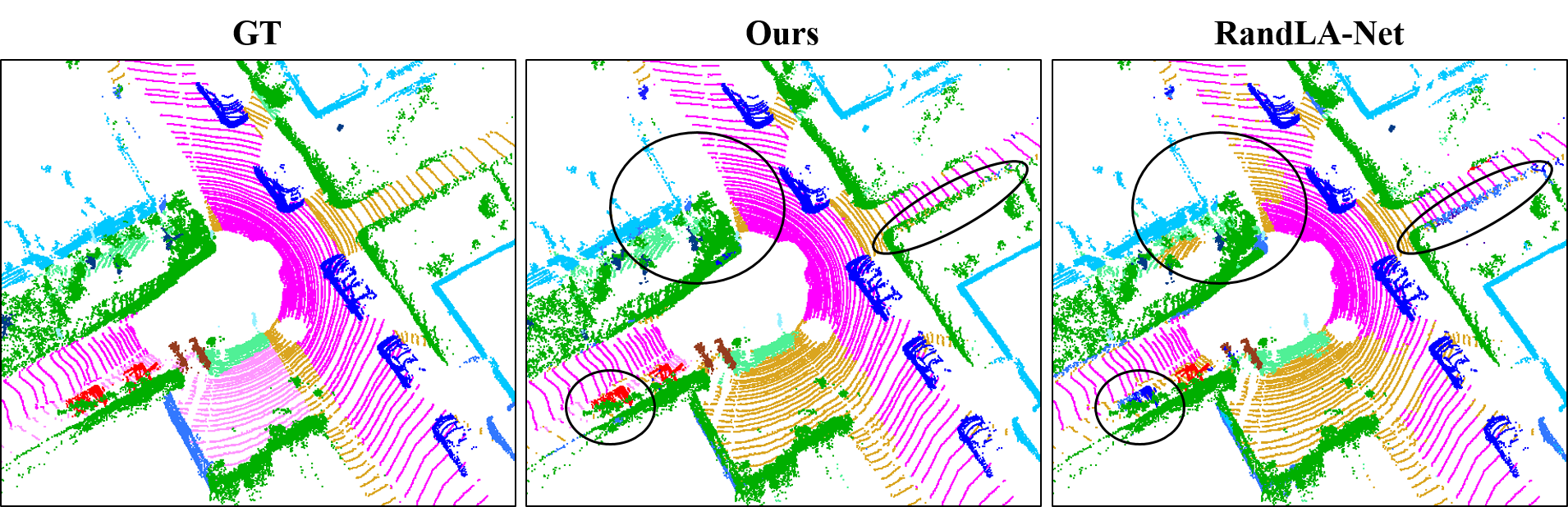}\\
		\caption{Semantic segmentation results of RandLA-Net~\cite{hu2019randla} and ours on the SemanticKITTI dataset. 
		}
		\label{semantickitti_compare}
	\end{minipage} 
	\begin{minipage}[b]{0.4\linewidth}
		\centering
		\tiny 
		\resizebox{1\textwidth}{!}{
			\begin{tabular}{lc}
				\toprule
				methods & mIoU(\%)\\
				\midrule
				Conv. on unordered points &64.15\\
				Isotropic Conv. after max-pooling &65.31\\
				Isotropic Conv. after mean-pooling &64.17\\
				Isotropic Conv. after sum-pooling &64.87\\
				Anisotropic Conv. (ours) &66.56\\
				\bottomrule
		\end{tabular}}
				\captionof{table}{Performance with several network variants on the S3DIS dataset.}
				\label{variants}
	\end{minipage}
\end{figure*}

\textbf{SemanticKITTI dataset}~\cite{behley2019semantickitti}: 
We finally compare the proposed STPC method with two families of recent point clouds semantic segmentation approaches, including point based methods~\cite{qi2017pointnet,landrieu2018large,qi2017pointnet++,tatarchenko2018tangent,hu2019randla} and projection based approaches~\cite{wu2018squeezeseg,wu2019squeezesegv2,behley2019semantickitti,milioto2019rangenet++}. 
The quantitative comparison of segmentation results is reported in Table~\ref{SemanticKITTI}. 
When compared with all point based methods and projection based approaches, the STPC can achieve the best performance in term of mIoU score, and also obtain the better results in most categories (11 out of 19). 
Especially in the motorcyclist category, the mIoU results of all these existing methods are quite low (e.g., 0.0\% with PointNet~\cite{qi2017pointnet}, SPLATNet~\cite{su2018splatnet} and PointNet++~\cite{qi2017pointnet++}, 0.9\% with SqueezeSeg~\cite{wu2018squeezeseg}, 7.2\% with RandLA-Net~\cite{hu2019randla}), but we can achieve 15.3\% segmentation result, which is much higher than the second-ranked TangentConv method~\cite{tatarchenko2018tangent} by improving 7.2\%. 
Compared with the best projection based approach 
(i.e., RangeNet53++~\cite{milioto2019rangenet++}), we outperform it by 2.4\% in term of  mIoU score, and obtain the better results in 14 categories.  Compared with the best point based method (i.e., RandLA-Net~\cite{hu2019randla}), we outperform it by 0.7\% in term of  mIoU score, and achieve the better results in 16 categories. 
Some visualization results between our STPC method and the compared RandLA-Net~\cite{hu2019randla} can be found in Fig.~\ref{semantickitti_compare}.
These above experimental results indicate that our proposed STPC method can better predict the point-wise category information of point clouds data by improving the predictive capability of the network.

\begin{table} \vspace{-0.4cm}
	\caption{Comparison of semantic segmentation performance on the SemanticKITTI dataset}
	\label{SemanticKITTI}
	\centering
	\resizebox{\textwidth}{!}{
		\begin{tabular}{rcccccccccccccccccccccc}
			\toprule
			Methods &Size &\rotatebox{90}{mIoU(\%)} &\rotatebox{90}{road} &\rotatebox{90}{sidewalk} &\rotatebox{90}{parking} &\rotatebox{90}{other-ground} &\rotatebox{90}{building} &\rotatebox{90}{car} &\rotatebox{90}{truck} &\rotatebox{90}{bicycle} &\rotatebox{90}{motorcycle} &\rotatebox{90}{other-vehicle} &\rotatebox{90}{vegetation} &\rotatebox{90}{trunk} &\rotatebox{90}{terrain} &\rotatebox{90}{person} &\rotatebox{90}{bicyclist} &\rotatebox{90}{motorcyclist} &\rotatebox{90}{fence} &\rotatebox{90}{pole} &\rotatebox{90}{traffic-sign} \\
			\midrule
			SqueezeSeg~\cite{wu2018squeezeseg} &\multirow{5}{1cm}{64*2048 \centering{pixels}} &29.5 &85.4 &54.3 &26.9 &4.5 &57.4 &68.8 &3.3 &16.0 &4.1 &3.6 &60.0 &24.3 &53.7 &12.9 &13.1 &0.9 &29.0 &17.5 &24.5\\
			SqueezeSegV2~\cite{wu2019squeezesegv2} & &39.7 &88.6 &67.6 &45.8 &17.7 &73.7 &81.8 &13.4 &18.5 &17.9 &14.0 &71.8 &35.8 &60.2 &20.1 &25.1 &3.9 &41.1 &20.2 &36.3\\
			DarkNet21Seg~\cite{behley2019semantickitti} & &47.4 &91.4 &74.0 &57.0 &26.4 &81.9 &85.4 &18.6 &26.2 &26.5 &15.6 &77.6 &48.4 &63.6 &31.8 &33.6 &4.0 &52.3 &36.0 &50.0\\
			DarkNet23Seg~\cite{behley2019semantickitti} & &49.9 &\textbf{91.8} &74.6 &64.8 &\textbf{27.9} &84.1 &86.4 &25.5 &24.5 &32.7 &22.6 &78.3 &50.1 &64.0 &36.2 &33.6 &4.7 &55.0 &38.9 &52.2\\
			RangeNet53++~\cite{milioto2019rangenet++} & &52.2 &\textbf{91.8} &\textbf{75.2} &\textbf{65.0} &27.8 &87.4 &91.4 &25.7 &25.7 &34.4 &23.0 &80.5 &55.1 &64.6 &38.3 &38.8 &4.8 &58.6 &47.9 &\textbf{55.9}\\
			\midrule
			PointNet~\cite{qi2017pointnet} &\multirow{6}*{50K pts} &14.6 &61.6 &35.7 &15.8 &1.4 &41.4 &46.3 &0.1 &1.3 &0.3 &0.8 &31.0 &4.6 &17.6 &0.2 &0.2 &0.0 &12.9 &2.4 &3.7\\
			SPGraph~\cite{landrieu2018large} & &17.4 &45.0 &28.5 &0.6 &0.6 &64.3 &49.3 &0.1 &0.2 &0.2 &0.8 &48.9 &27.2 &24.6 &0.3 &2.7 &0.1 &20.8 &15.9 &0.8\\
			SPLATNet~\cite{su2018splatnet} & &18.4 &64.6 &39.1 &0.4 &0.0 &58.3 &58.2 &0.0 &0.0 &0.0 &0.0 &71.1 &9.9 &19.3 &0.0 &0.0 &0.0 &23.1 &5.6 &0.0\\
			PointNet++~\cite{qi2017pointnet++} & &20.1 &72.0 &41.8 &18.7 &5.6 &62.3 &53.7 &0.9 &1.9 &0.2 &0.2 &46.5 &13.8 &30.0 &0.9 &1.0 &0.0 &16.9 &6.0 &8.9\\
			TangentConv~\cite{tatarchenko2018tangent} & &40.9 &83.9 &63.9 &33.4 &15.4 &83.4 &90.8 &15.2 &2.7 &16.5 &12.1 &79.5 &49.3 &58.1 &23.0 &28.4 &8.1 &49.0 &35.8 &28.5\\
			RandLA-Net~\cite{hu2019randla} & &53.9 &90.7 &73.7 &60.3 &20.4 &86.9 &94.2 &\textbf{40.1} &26.0 &25.8 &\textbf{38.9} &81.4 &61.3 &66.8 &49.2 &48.2 &7.2 &56.3 &49.2 &47.7\\
			\midrule
			Our STPC &50K pts &\textbf{54.6} &90.8 &74.1 &63.6 &5.3 &\textbf{90.7} &\textbf{94.7} &34.4 &\textbf{48.9} &\textbf{39.7} &24.5 &\textbf{82.7} &\textbf{62.1} &\textbf{67.5} &\textbf{51.1} &\textbf{48.9} &\textbf{15.3} &\textbf{61.5} &\textbf{51.4} &47.9\\
			\bottomrule
	\end{tabular}} \vspace{-0.4cm}
\end{table}
\begin{figure*}[!t]
\begin{minipage}[t]{0.5\linewidth}
  \centering
  \includegraphics[width=6.5cm]{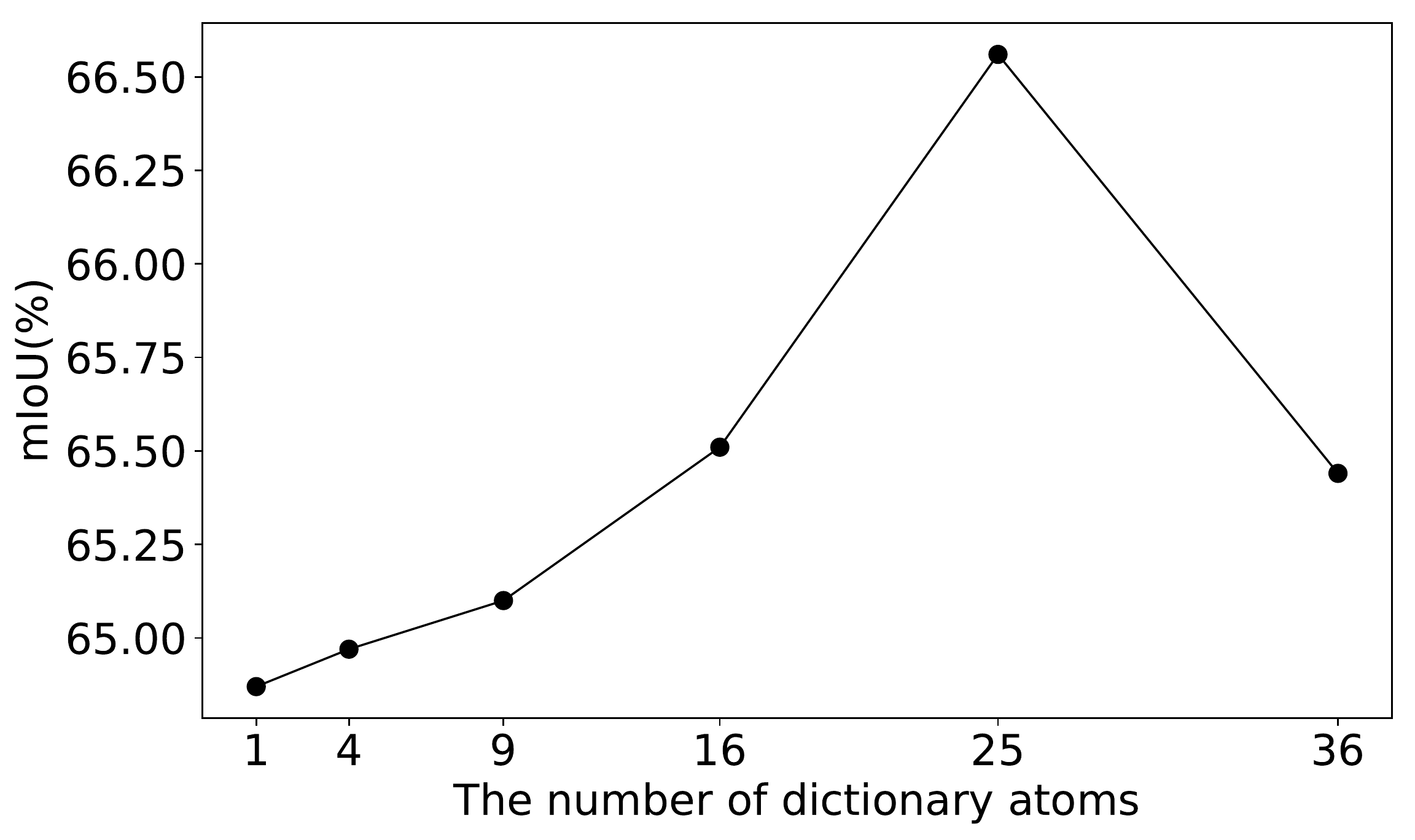}\\
  \caption{Segmentation results with different number of dictionary atoms on the S3DIS dataset.}
  \label{change atoms number}
\end{minipage} \hspace{0.2cm}
\begin{minipage}[t]{0.5\linewidth}
  \centering
  \includegraphics[width=6.5cm]{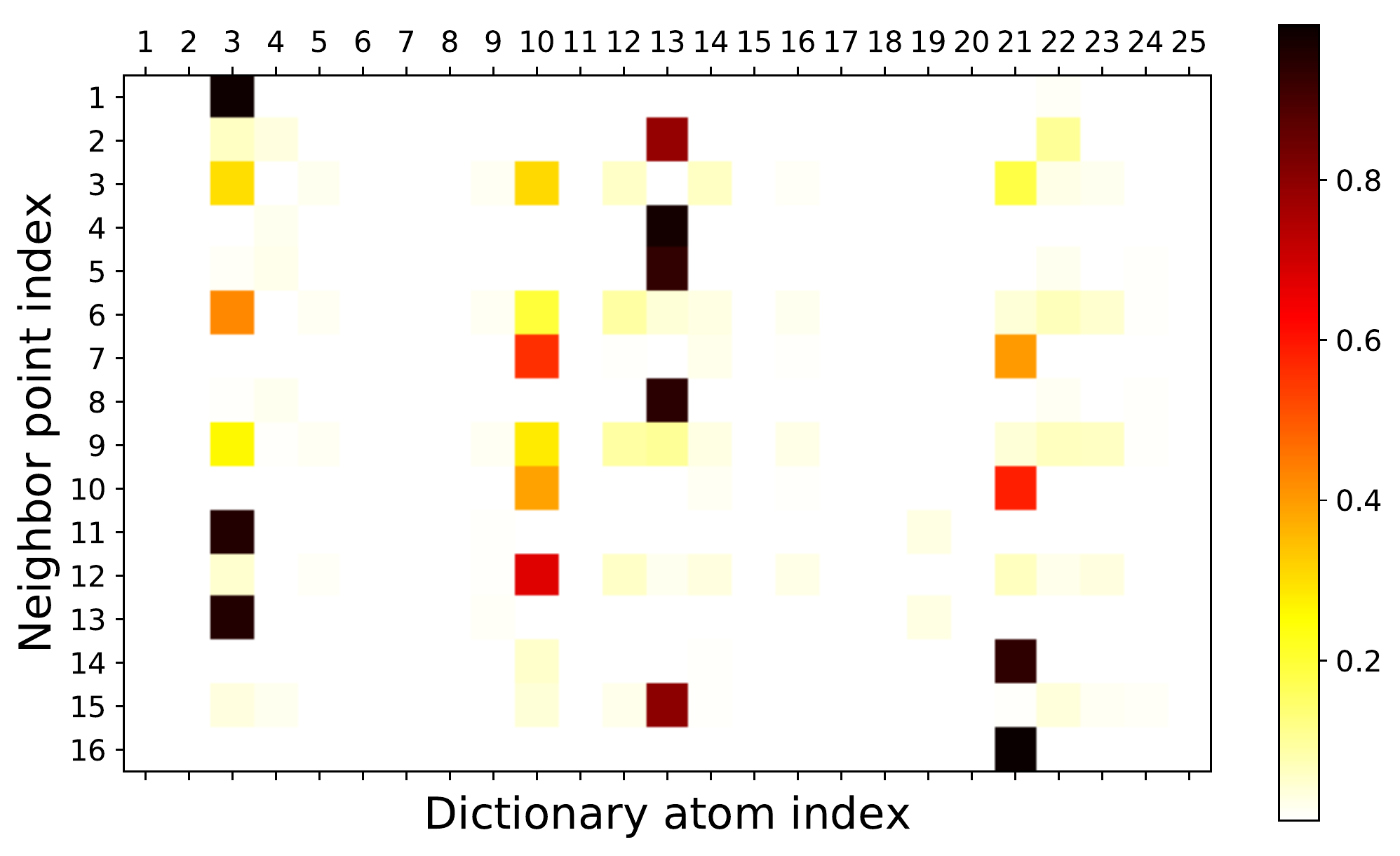} 
  \caption{The direction-dependent relationship between one neighbor point $p_k$ and each atom $\a_m$.}
  \label{attention0} \vspace{-0.2cm}
\end{minipage} 
\end{figure*}
\vspace{-0.2cm}
\subsection{Ablation Study}\vspace{-0.2cm}
In this section, we conduct the following ablation studies for our spatial transformer point convolution module. All ablated networks are trained on area 1$\sim$4 and 6, and tested on area 5 of S3DIS dataset.

As illustrated in Figure.\ref{change atoms number}, we compare the semantic segmentation performance with different number of atoms (i.e., $M=1, 4, 9, 16, 25, 36$) in the spatial direction dictionary.
When the number of atoms increases from 1 to 25, the performance of point clouds semantic segmentation is also improving from 64.87\% to 66.56\% in term of mIoU score.  
If the number of atoms continues to increase, the segmentation performance would be slightly decline.
Our STPC method achieves the best result when the number of dictionary atoms is 25 ($M=25$). 
If the number of atoms is too small, the local geometric information cannot be captured finely in the anisotropic convolution filtering process. 
When the number of atoms equals to 1, our method would degenerate to the sum operation, while  
if the number of atoms is too large, the learnt weights tend to average on each atom, and these finer features would be lost. 
Therefore, a suitable number of dictionary atoms can make the proposed SPTC method to capture the feature varieties of point clouds. 



For better understanding the direction dictionary learning process, 
we visualize the learned representation coefficients in  Eqn.~\ref{eq:eps8}, which can reflect the direction-dependent relationship between one neighbor point $p_k$ and each atom $\a_m \in \A$. As shown in Fig.\ref{attention0}, the vertical ordinates refer to the index of the K nearest neighbor points (i.e., $p_k \in \mathcal{N}(p_i)$, $k=1, 2, \dots, 16$) and the horizontal  ordinates denote the index of dictionary atoms (i.e., $\a_m \in \A$,  $m=1, 2, \dots, 25$). 
The local neighbor points around the point $p_i$ can be projected onto different dictionary atoms.
It indicates that the feature-dense local point clouds can be sparsely encoded by sparse discrete direction dictionary.


For exploring the effectiveness of the anisotropic filtering/convolution in our network,  we evaluate the segmentation performance with several network variants on the S3DIS dataset. 
Specifically, we directly perform convolution on unordered point clouds (``Conv. on unordered points"), and conduct isotropic convolution by aggregating local spatial points after max-, mean-, and sum-pooling operations, named ``Isotropic Conv. after max-pooling", ``Isotropic Conv. after mean-pooling", and ``Isotropic Conv. after sum-pooling", respectively. 
As can be seen in Table~\ref{variants}, when compared with ``Conv. on unordered points", our proposed anisotropic filtering can improve the segmentation perform from 64.15\% to 66.56\%. 
When compared with three different isotropic convolutions, 
the performance of our proposed anisotropic convolution is better than 
three isotropic convolution methods, representing improvements of 1.25\%, 2.39\% and 1.69\% over ``Isotropic Conv. after max-pooling", ``Isotropic Conv. after mean-pooling", and ``Isotropic Conv. after sum-pooling". 
It proves that our anisotropic filtering method (i.e., STPC) can capture and represent implicit geometric structures, which is more robust to express those finer variances of local regions.

\section{Conclusion}
In this work, we track the point clouds semantic segmentation problem with the spatial transformer point convolution method, which can better predict the point-wise category information of 3D point clouds. 
By employing the spatial direction information of point clouds, we learn the spatial direction dictionary to represent those latent geometric components. 
By projecting these unordered neighbor points into the canonical dictionary space, we introduce the spatial transformer point convolution to perform the anisotropic filtering process.
The direction dictionary learning, spatial direction encoding, spatial anisotropic filtering processes can be integrated into an unified network and jointly optimized in an end-to-end fashion. 
Extensive experimental results clearly demonstrate the effectiveness of the proposed spatial transformer point convolution method.

\clearpage
\textbf{\large Broader Impact}

\textbf{Future Societal Consequences.} 
Recent advances in 3D hardware sensors (e.g., 3D scanners, LiDARs and RGB-D cameras) has improved the accessibility of point clouds, which can intuitively reflect the spatial geometric information of objects in 3D space. 
3D semantic segmentation of point clouds, which aims to label each point with a corresponding category information, has drawn increasing attention due to its broad applicability to the fields of general environment perception. 
It enables various higher-level potential applications, including autonomous driving, human-computer interaction, virtual reality and robotics. 
Meanwhile, with the development of artificial intelligence, deep learning has made significantly progress in the field of computer vision. 

This work "Spatial Transformer Point Convolution" focuses on the anisotropic convolution filtering method on point clouds. 
Different from these previous methods, our proposed STPC can perform the anisotropic convolution filtering on point clouds by expressing those finer variances of local regions, and then further improve the performance of point clouds semantic segmentation. Extensive experiments have proved its discriminative capability. 

This work will be of great significance for the applications in the 3D semantic segmentation domain and will further promote the development of artificial intelligence.

\textbf{Ethical Consideration.}
This work mainly performs scene understating in 3D space, and facilitate the
development of some high-level vision applications. For example, it can help understand our living environment to facilitate the human-computer interaction.  For the negative outcome, it will depend on the specific task and the criteria for assessing positive and negative. Besides, in this paper, all used datasets are publically available as academic research, and the evaluation metrics are also standard.

\clearpage

\small
\bibliographystyle{plain}
\bibliography{neurips_2020}

\end{document}